\documentclass[letterpaper, 10 pt, conference]{ieeeconf}  

\IEEEoverridecommandlockouts                              

\usepackage{graphics} 
\usepackage[T1]{fontenc}
\usepackage{epsfig} 
\usepackage{mathptmx} 
\usepackage{times} 
\usepackage{amsmath} 
\usepackage{amssymb}  
\DeclareMathAlphabet{\altmathcal}{OMS}{cmsy}{m}{n}  
\usepackage{algorithm}
\usepackage{algpseudocode}
\usepackage{subcaption}

\title{\LARGE \bf
Automatically Learning Fallback Strategies with Model-Free Reinforcement Learning in Safety-Critical Driving Scenarios
}

\author{Ugo Lecerf$^{1,2}$, Christelle Yemdji Tchassi$^1$, S\'{e}bastien Aubert$^1$, Pietro Michiardi$^2$
\thanks{$^{1}$Renault Software Labs, Valbonne, France
        {\tt\small \{ugo.lecerf, christelle.yemdji-tchassi, sebastien.s.aubert\}@renault.com}}%
\thanks{$^{2}$Eurecom, Biot, France.
        {\tt\small \{ugo.lecerf, pietro.michiardi\}@eurecom.fr}}%
\thanks{\hrule
Published in Proceedings of 7$^{th}$ edition of International Conference of Machine Learning Technologies (ICMLT) 2022}%
}

\begin{document}

\maketitle
\thispagestyle{empty}
\pagestyle{empty}

\begin{abstract}

When learning to behave in a stochastic environment where safety is critical, such as driving a vehicle in traffic, it is natural for human drivers to plan fallback strategies as a backup to use if ever there is an unexpected change in the environment. Knowing to expect the unexpected, and planning for such outcomes, increases our capability for being robust to unseen scenarios and may help prevent catastrophic failures. Control of Autonomous Vehicles (AVs) has a particular interest in knowing when and how to use fallback strategies in the interest of safety. Due to imperfect information available to an AV about its environment, it is important to have alternate strategies at the ready which might not have been deduced from the original training data distribution. 

In this paper we present a principled approach for a model-free Reinforcement Learning (RL) agent to capture multiple modes of behaviour in an environment. We introduce an extra pseudo-reward term to the reward model, to encourage exploration to areas of state-space different from areas privileged by the optimal policy. We base this reward term on a distance metric between the trajectories of agents, in order to force policies to focus on different areas of state-space than the initial exploring agent. Throughout the paper, we refer to this particular training paradigm as learning fallback strategies.

We apply this method to an autonomous driving scenario, and show that we are able to learn useful policies that would have otherwise been missed out on during training, and unavailable to use when executing the control algorithm.

\end{abstract}

\section{INTRODUCTION}
Implementing a controller for AVs in a driving scenario is met with many challenges: both from the point of view of perception and control \cite{AD_Survey_common_practices2020}. As in most applications of real-world RL, the uncertainty linked to the perception of the agent's environment must be considered for an effective controller to be developed. Even with the best possible road maps and sensors, it is impossible to eliminate all sources of uncertainty from a driving scenario, be they epistemic from imperfections in the vehicle's sensors, or aleatoric from the unpredictable interactions with other drivers \cite{depeweg2018decomposition}. 

Autonomous driving requires a strong notion of safety, and notably robustness with respect to unexpected changes in the agent's environment. For example, sensor perception quality can be heavily susceptible to adverse weather conditions \cite{Impact_Adverse_weather_conditions_on_sensors}. Because of this the optimal behaviour is likely to change dynamically according to the vehicle's inputs, and a satisfactory control algorithm must be able to adapt on the fly. Safety criteria in autonomous driving applications are traditionally based on perceiving when a situation is no longer able to be handled by the acting controller, and then handing over the controls to either the driver, or a special-case controller. For example, \cite{Bouton19_scene_decomposition} implement a deep neural network to detect the probability of a catastrophic outcome when following recommended actions, whereas other approaches look to estimate the confidence of a policy by estimating quantiles of the return distribution \cite{clements2019estimating}, or using an ensemble of networks to gauge whether an input was included in the policy's training data distribution \cite{hoel20_IV_ensemble_uncertainty}.

RL techniques have been shown to be able to tackle the task of control in progressively more complex environments \cite{badia2020agent57}. RL algorithms learn by optimizing their expectation of performance in an environment. In most cases, such as board games \cite{mnih2015} or video games \cite{openai2019dota}, the environment in which we seek to obtain the optimal behaviour can be modeled as a Markov Decision Process (MDP) with no loss of generality in the solution found by the RL agent. Through advancements in target updates \cite{hessel2017_rainbowDQN}, as well as agent architectures \cite{schaul2015prioritized}, RL agents have become increasingly efficient at finding the optimal solution to MDPs, even when requiring high degrees of exploration, where the optimal sequence of actions is hard to find \cite{SuttonF1998PracticalRL}. 

Stochastic control environments such as driving scenarios are more difficult to optimize, given the probabilistic nature of both the observation and transition dynamics. Stochastic environments may be modeled as Partially Observable MDPs (POMDPs) \cite{Kochenderfer_book}. Solving POMDPs is possible with methods combining learning and planning, such as \cite{Hoel2019}. However, a change in the values of the stochastic model parameters, for example a change in a vehicle's sensor accuracy, scene obstruction, or simply unplanned behaviour from another vehicle, may induce a sharp drop in the agent's performance due to its inability to generalize well to new environment parameters. Having access to a model of the environment dynamics allows us to use planning algorithms, such as MCTS \cite{mcts_survey}, alongside learning to both increase sample efficiency, and have access to a better representation of the environment's state-space structure (model-based RL). In cases where planning is possible, it is much easier to find alternative strategies for an agent to solve its environment and hence be able to better adapt to eventual changes in the state-space \cite{mcallister_pilco_for_pomdps}.

In the model-free setting we consider in this paper however, efficient learning is notoriously hard, one of the factors being the increased variance of target updates from of the lack of a planner able to average out return values from multiple simulated runs. A lack of an environment model also reduces the ability for an agent to adapt to changes to the environment after training, since we are unable to use a planner to explore the new dynamics before acting. Works such as \cite{Hausknecht_DRQL} and \cite{Zhu_DRL_POMDP} use deep recurrent $Q$-networks in order to build up knowledge of the MDP's state over time, and use a latent representation of the history of states and actions in order to inform an agent about its current state. In the case of stochastic parameter change, we are unable to predict how the returns of a policy will be affected. 

This work is focused on the problem of training an agent to be robust to changes in the uncertainty affecting local areas of state-space affecting either the transition or observation models in the MDP-defined environment. When uncertainty arises in a local area of an MDP's state-space, we can try to learn policies exploiting different areas of the state-space, such that we maximize the probability that the return of at least one of the available policies remains unaffected.

Our main contribution is to implement a novel framework which aims to learn alternate policies, referred to as fallback strategies, which exploit different areas of state-space than the optimal policy. These fallback strategies serve as potential alternatives to safely navigate the environment, in the case of a change in the defining elements in the MDP model. We consider the model-free RL setting, and provide experimental validation for this framework by testing it in a driving scenario.

This paper is organized as follows: in section \ref{sec:notation} we present the standard framework and notations used in RL. Section \ref{sec:related_works} presents existing methods for affecting agent behaviour in RL systems. Section \ref{sec:fallback_strategies} presents our contribution of learning fallback strategies, the main topic of this paper, where we define multi-policy learning in the context of learning and retaining sub-optimal solutions. Experiments and results are presented in sections \ref{sec:experiments} and \ref{sec:results} respectively, and section \ref{sec:conclusion} contains our conclusion along with future work.

\section{NOTATION}
\label{sec:notation}
A finite MDP is defined by the following elements: 
\begin{itemize}
    \item Finite set of states $s\in\altmathcal{S}$. States are indexed by the timestep at which they are encountered: $s_t$.
    \item Finite set of actions $a\in \altmathcal{A}$. Actions are also indexed by their respective timesteps: $a_t$.
    \item Transition model $\altmathcal{T}(s,a,s'):\altmathcal{S}\times\altmathcal{A}\times\altmathcal{S} \rightarrow [0,1]$, representing the probability of passing from $s$ to $s'$ after taking action $a_t$, $P(s_{t+1}=s'|s_t=s, a_t=a)$.
    \item Immediate reward function $R(s,a,s'):\altmathcal{S} \times \altmathcal{A}\times\altmathcal{S} \rightarrow \mathbb{R}$.
    \item Discount factor $\gamma\in (0,1]$, controlling the weight in value of states further along the Markov chain.
\end{itemize}
A POMDP is further augmented by an observation model $\altmathcal{O}$, when we no longer have access to the true state $s_t$, but an observation thereof $o_t=\altmathcal{O}(s_t)$.

Actions in the MDP are taken by a (stationary) policy $\pi :\altmathcal{S} \rightarrow \altmathcal{A}$ mapping states to actions. The value of a state under a policy $\pi$, is given by the state-value function $V^\pi:\altmathcal{S}\rightarrow \mathbb{R}$, which represents the expected future discounted sum of rewards, if policy $\pi$ is followed from $s$:
\begin{gather}
\label{eq:value_fn_definition}
V^\pi(s):=\mathbb{E}\left[\sum^\infty_{t=0}\gamma^t R(s_t,a_t,s_{t+1}) \right], \\ 
s_0=s, a_t=\pi(s_t), s_{t+1}\sim\altmathcal{T}(s_t,a_t). \nonumber
\end{gather}
Actions are chosen by the policy $\pi$, so as to maximize the action-value function $Q^\pi:\altmathcal{S}\times\altmathcal{A} \rightarrow \mathbb{R}$ which assigns values to actions according to the value of the states that are reached:
\begin{gather}
\label{eq:q_fn}
    Q^\pi(s_t,a_t):=\mathbb{E}_{s_{t+1}}\left[R(s_t,a_t,s_{t+1})+\gamma\cdot V^\pi(s_{t+1})\right], \\ 
    s_{t+1}\sim \altmathcal{T}(s_t,a_t). \nonumber
\end{gather}

We use the $Q$-function in order to define the optimal policy, which we denote $\pi^*$, as the policy taking actions that maximize the action-value function: $\pi^*(s):=\text{arg}\max_{a\in\altmathcal{A}}Q^{\pi^*}(s,a)$. Equation (\ref{eq:q_fn}) highlights that the state-action value function is a sort of one-step look-ahead of the value of the next possible state $s_{t+1}$, in order to determine the value of actions in the current state $s_t$. We denote the history of states visited by a policy $\pi$ (trajectory through state-space) as: 
\begin{gather}
\altmathcal{H}_\pi:=\left\{s_t\right\}_{t\in[0,T]}, \\
s_0\in\altmathcal{S}, a_t=\pi(s_t), s_{t+1}\sim\altmathcal{T}(s_t,a_t). \nonumber
\end{gather}

\section{RELATED WORK}
\label{sec:related_works}

The main element susceptible to engineering in MDPs is the reward function, which indirectly defines the agent's goal; learning about multiple goals can be translated into learning to solve MDPs with multiple reward functions. One of the most common uses for augmenting the reward function is to artificially boost the RL agent's degree of exploration \cite{NeverGiveUp_Badia_2020}, \cite{eriksson2019epistemic}. These methods dynamically change the value of the immediate rewards an agent gains, in order to encourage actions towards areas of state-space which are deemed more important. Our approach is similar to these in that we base an extra reward term on an external factor in order to affect the behaviour of exploring agents. 

Although we base our approach on engineering the MDP's reward function, the problem we are tackling with our approach is distinct from the exploration problem in RL, and our objective is not to find an optimal exploration scheme. Exploration boosting methods are used within the context of a single environment, in order to avoid having the agent fall victim to being trapped within a local optima. Our goal however is to have the RL algorithm be able to retain sub-optimal solutions to the environment, once the training regime is ended. In a similar spirit to how TRPO \cite{schulman2015trpo} seeks to mitigate the concern that small changes in the parameter space may lead to sharp drops in performance, we seek to mitigate sharp drops in performance resulting from changes to stochastic environment parameters.

Moreover, the approach explored in this paper aims to learn policies with different behaviours, given the same initial environment conditions. Compared to methods which learn only the environment's optimal policy, this allows us to have sub-optimal policies to use as fallback strategies which we can switch through using a hierarchical structure. This has an advantage over attempting to have a single policy learn to generalize over the space of MDPs, since this task (meta-learning) requires many more samples in order for an agent to achieve a good performance \cite{kirsch2019metagenRL}, our approach is more attractive for certain safety-critical applications.

Many previous works aim for agents to generalize to different goals within the same MDP \cite{UVFA_Schaul_2015}, or even self-discover sub-goals that make learning more efficient \cite{Machado_LaplacianOptionDiscovry}. In these cases, a difference in goals is translated through a change in the reward function for a goal-state $g$: $R(s,a,g)=R_g>0$. These methods take advantage of the underlying structure in the goal-space in order to increase sample efficiency \cite{HER_Anndrychowicz_2018}, as well as the generalization capabilities of agents to tasks with goals that were not present during training \cite{Eysenbach_search_on_the_replay_buffer}. They provide a good way of finding different behaviours within a same environment, where the modification of the reward function is intended for a control algorithm to be robust in terms of changing objectives. Our problem statement focuses on dealing with scenarios when the objective (i.e. goal) of an environment remains the same, but the environment is expected to change in some unknown way. 

\section{LEARNING FALLBACK STRATEGIES}
\label{sec:fallback_strategies}
In this section, we motivate the use of a new RL paradigm aimed at learning policies in the model-free setting which are robust to changes in the environment's stochastic parameters, affecting its dynamics locally. Note that because of the lack of access to environment models, we can only target the sampled transitions that are stored in the replay buffer used during training.

\subsection{Agents' Behaviour}
We can think of an agent's behaviour as the trajectory through state-action-space that results from following a policy $\pi$. Policies with similar behaviours (condition which we formally define later on) will have similar trajectories through state-action-space. During training while an agent's policy is improving, the agent's behaviour will evolve, until its optimal policy is reached.

When considering learning multiple policies in an environment, we must determine which ones are useful to us. We consider that any policy which sufficiently solves an MDP is of interest. \emph{Sufficiently solved} is a criterion that may vary between different MDPs, hence we identify two ways this can be implemented into RL algorithms:

\vspace{2mm}
\noindent \textbf{Definition} (Sufficiently Solved). An agent's policy $\pi$ sufficiently solves its respective MDP, if either of the following conditions are met, depending on the nature of the control task:
\begin{itemize}
    \item Agent is able to reach a specific goal-state $g\in\altmathcal{S}$.
    \item Agent is able to expect accumulated discounted rewards above a threshold score $V^{\pi}(s_0)>G_{min}\in\mathbb{R}$, for $s_0$ sampled from initial state distribution (Note that $V^\pi$ is the true value function as in (\ref{eq:value_fn_definition}), not an approximation).
\end{itemize}

Based on this, we define what we mean by a valid strategy:

\vspace{2mm}
\noindent \textbf{Definition} (Valid Strategy). A valid strategy in an MDP, is the behaviour of a policy which sufficiently solves that MDP.

\vspace{2mm}
The use of either interpretation for \emph{sufficiently solved} depends on the environment, and what it heuristically means to solve it. For example, in the case of an Atari game (e.g. Breakout), any behaviour from a policy which reaches above a threshold score, is typically considered to be a valid strategy. Another example would be an AV passing through an intersection, where any behaviour passing the intersection (reaching goal-state $g$) without collisions is a valid strategy. We use this definition to determine which policies are deemed useful during training. 

\vspace{2mm}
\noindent \textbf{Definition} (Sub-optimal policy). A sub-optimal policy, denoted $\pi_{sub}$, is a policy whose expectation of return is within a margin $\epsilon$, to that of the optimal policy $\pi^*$, at some given initial state $s_0$ sampled from the initial state distribution:

\begin{equation}
\label{eq:value_fn}
    V^{\pi^*}(s_0)-V^{\pi_\text{sub}}(s_0)<\epsilon.
\end{equation}

Condition (\ref{eq:value_fn}) is equivalent to saying that policy $\pi_\text{sub}$ is a sub-optimal policy, whose behaviour is a valid strategy (in the threshold-score sense, $\epsilon=V^{\pi^*}(s_0)-G_{min}$). For example in the case of an Atari game, any agent that achieves a score higher than the threshold, but less than the one obtained by the optimal policy $\pi^*$, verifies condition (\ref{eq:value_fn}). 

In order for the two policies to be considered as having different behaviours, they must be sufficiently different in the state-distributions that are encountered during execution. This implies the need for a metric $\altmathcal{M}$ measuring the difference between agents' trajectories in state-space, which is not yet a part of standard reinforcement learning applications.

\vspace{2mm}
\noindent\textbf{Definition} (Sufficiently different behaviours). We can say that two policies, $\pi_1$, $\pi_2$ have $\altmathcal{M}_d$-different behaviours, iff:
\begin{equation}
\label{eq:state_distrib}
    \altmathcal{M}\left(
    \mathbb{E}\left[\altmathcal{H}_{\pi_1}\right],  \mathbb{E}\left[\altmathcal{H}_{\pi_2}\right] \right) \geq d.
\end{equation}

Condition (\ref{eq:state_distrib}) is equivalent to saying that the behaviours of $\pi_1$ and $\pi_2$ can be described as being heuristically different. In our approach, we base this heuristic on the similarity in terms of their respective trajectories through the MDP state-space. Setting a value $d\in\mathbb{R}$ is subjective: for instance, human experts may have arbitrary boundaries for when an agent's path through state-space is sufficiently different from a reference path, to consider both as having different behaviours. Condition (\ref{eq:state_distrib}) can be thought of as a non-parametric clustering with boundary $d$, where $\altmathcal{M}(\cdot,\mathbb{E}[\altmathcal{H}_{\pi_{ref}}])$ is the feature map in a policy's state-trajectory space, with respect to a reference policy $\pi_{ref}$. Once more, the correct segmentation of this space is subjective and may vary between experts based on experience. 

We propose that in order to increase the robustness of an RL algorithm to changes in environment parameters, it should be able to learn sub-optimal strategies which have sufficiently different behaviours from each other policy in a training environment, satisfying both conditions (\ref{eq:value_fn}) and (\ref{eq:state_distrib}). Changes in $\altmathcal{T}$ or $\altmathcal{O}$ may affect local areas of the environment's state-space differently, affecting some policies' expected returns more than others, depending on whether the introduced uncertainty affects their respective state-space paths. Condition (\ref{eq:value_fn}) ensures that we only learn policies with satisfactory performance in the environment, whereas (\ref{eq:state_distrib}) aims to maximize the likelihood that at least one of the learned policies will have an expected return that is minimally affected by changes to either $\altmathcal{T}$ or $\altmathcal{O}$.

\subsection{Fallback Strategies}

\begin{figure}
\vspace*{2mm}
    \centering
    \includegraphics[width=8cm]{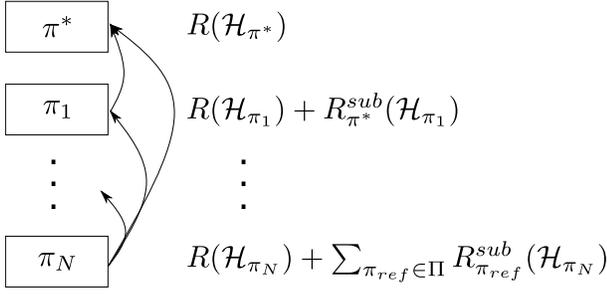}
    \caption{Each subsequent pseudo-agent will have its relative pseudo-reward added onto its regular reward function, each extra term corresponding to another agent already present in that environment. This way, we verify that new agents have behaviours that are sufficiently different from those of all previous agents' policies.}
    \label{fig:N_agents}
\end{figure}

Given an MDP, we wish to find the optimal policy along with a number $N$ of sub-optimal ones. We consider situations where the agent is already able to find the optimal policy $\pi^*$, and demonstrate a method for finding policies $\pi_{sub}$ that satisfy both (\ref{eq:value_fn}) and (\ref{eq:state_distrib}).

The approach explored in this paper is to add an extra pseudo-reward term onto the regular reward for that environment, denoted $R^{sub}_{\pi_{ref}}\leq 0$, as a penalty to agents, for having similar state-trajectories to $\pi_{ref}$. We term the agents which are not seeking to learn the optimal policy, pseudo-agents. The pseudo-reward is based on the metric $\altmathcal{M}$ between agents' state-space trajectories in order to satisfy (\ref{eq:state_distrib}). We use the following equation for the pseudo-reward to discourage pseudo-agents agents from copying other agents' expected path through state-space $\mathbb{E}\left[\altmathcal{H}_{\pi_{ref}}\right]$:

\begin{equation}
R^{sub}_{\pi_{ref}}(\altmathcal{H}_\pi)= -\frac{\alpha}{\altmathcal{M}\left(
    \altmathcal{H}_\pi, \mathbb{E}\left[\altmathcal{H}_{\pi_{ref}}\right]\right) + \delta} ,
\label{eq:neg_rwd_general_formula}
\end{equation}
where $0<\delta<1$ avoids infinite penalties for exactly following  $\mathbb{E}\left[\altmathcal{H}_{\pi_{ref}}\right]$, giving the penalty an upper bound of $-\frac{\alpha}{\delta}$. $\pi_{ref}$ is the reference policy, which may or may not be the optimal, according to the number of pseudo-agents. $\alpha$ is a scaling factor to adjust the amplitude of the pseudo-reward term, compared to the regular rewards. Pseudo-agents will be training concurrently to the optimal one, aiming to converge to distinct valid strategies within the same environment. 

Fig. \ref{fig:N_agents} illustrates the relationship between subsequent pseudo agents in the same environment.  This approach can be extended to an arbitrary number $N$ of pseudo-agents, by imposing condition (\ref{eq:state_distrib}) such that each subsequent agent has a sufficiently different behaviour to previous ones, hence every additional pseudo-agent will have one more reward term to compute. Although this adds complexity to the RL problem, the number $N$ of total agents should remain reasonably limited: we should increase $N$ according to the anticipated uncertainty on the environment parameters. $\pi_{ref}\in\Pi$ represents all previous agents (shown by the arrows in Fig. \ref{fig:N_agents}). $\Pi$ is empty in the case of the optimal agent $\pi^*$, $\Pi=\{\pi^*\}$ for the 1st pseudo-agent, $\Pi=\{\pi^*, \pi_1\}$ for the 2nd pseudo-agent, and so on. For $N$ pseudo-agents, this approach adds a computational cost of $o(N^2)$ in terms of pseudo-reward term computation. Increasing state-space size and dimensionality is susceptible to increase the number $N$ of sub-optimal agents we wish to maintain as there are more opportunities for alternate valid strategies. However, $N$ is limited by the number of expected changes in the MDP's state-space we wish the RL agent be robust to, hence the computational cost is expected to remain within the same order of magnitude as without the pseudo-reward implementation.

Implementing the additional pseudo-rewards will impact the new value-function estimate of pseudo-agents learning sub-optimal policies . So (\ref{eq:value_fn}) should become:

\begin{equation}
 V^{\pi^*}(s_0)-V^{\pi_\text{sub}}(s_0)<\epsilon + \sum_{\pi_{ref}\in\Pi}R^{sub}_{\pi_{ref}}\left(\altmathcal{H}_{\pi_{sub}}\right),
 \label{eq:adjusted_value_fn}
\end{equation}
such that $\pi_{sub}$ would still be considered a valid sub-optimal policy.

Algorithm \ref{alg:concurrent} gives a pseudo-code description of our implementation. $\pi_{ref}\in\Pi$ represents the same set of reference agents as in Fig. \ref{fig:N_agents}.

\begin{algorithm}
\caption{Learning Fallback Strategies with N pseudo-agents}
\label{alg:concurrent}
\begin{algorithmic}[1]
\State Init $\pi^*, \pi_1, ..., \pi_N$
\While{True}
    \For{$\pi\in\{\pi^*, \pi_1, ..., \pi_N\}$} 
        \While{episode not terminated} \Comment{play episode}
            \State $a_t=\pi(s_t)$
            \State $s_{t+1}\sim\altmathcal{T}(s_t,a_t)$
            \State $r_t=R(s_t,a_t,s_{t+1})$ \Comment{regular step-reward}
        \EndWhile
        \State $r_{pseudo}=\sum_{\pi_{ref}\in\Pi}R^{sub}_{\pi_{ref}}\left(\altmathcal{H}_\pi\right)$ \Comment{pseudo-reward}
        \For{$t\in [0,T-2]$} \Comment{store in memory}
            \State Memory$\left(\pi\right)\gets \left(s_t,a_t,r_t,s_{t+1}\right)$
        \EndFor
        \State Memory$\left(\pi\right)\gets \left(s_{T-1},a_{T-1},r_{T-1}+r_{pseudo},s_T\right)$
    \EndFor
\EndWhile
\end{algorithmic}
\end{algorithm}

\section{EXPERIMENTS}
\label{sec:experiments}
In this section we present a control task for an autonomous vehicle, to demonstrate the ability of our proposed method to discover and maintain valid sub-optimal policies. In this use-case, we limit ourselves to a single fallback strategy.

\begin{figure}[tpb]
  \centering
  \includegraphics[width=8cm]{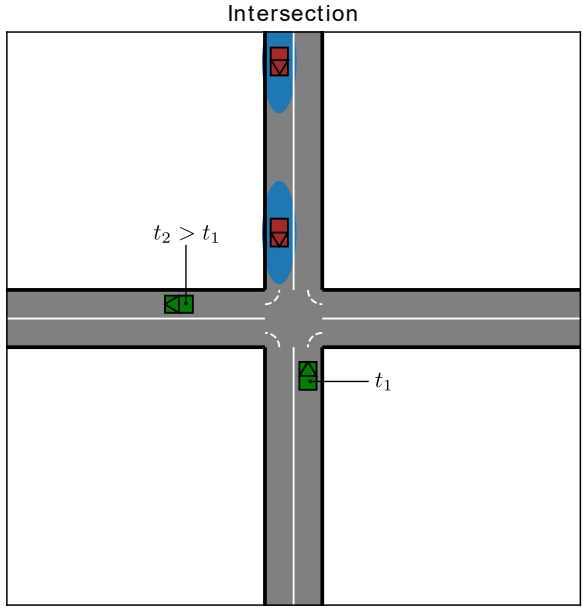}
  \caption{Intersection environment. The ego vehicle (green) starts on the bottom-right lane, and makes a left turn. The ego vehicle can either speed up to pass in front of the oncoming target vehicles, or slow down to pass in-between them. The aim is to cross the intersection without crashing ($t_2$). Collision distance is given by the radius of the blue ellipses.}
  \label{fig:mdp_cross_capture}
\end{figure}

The environment consists of a 2-way intersection, where the agent's goal is to complete a left-hand turn, without crashing into any of the oncoming vehicles that cross straight through the intersection. Fig. \ref{fig:mdp_cross_capture} shows a frame of the environment with oncoming vehicles in the intersection. In driving scenarios, the controllable agent is usually referred to as the ego vehicle whereas the other uncontrollable vehicles are referred to as targets. In this scenario there are two main solutions for the ego to complete the task: the optimal policy in terms of performance is to speed up and pass before the first target vehicle, whereas the sub-optimal policy consists of slowing down and passing in-between the oncoming target vehicles. A change in scene detection may affect the variance in detected positions of the target vehicles, and cause the first strategy to be considered too risky to follow. In this case, learning a fallback strategy that may be less affected by a drop in target position confidence, may be considered safer and more useful.

In our use-case, the speed of target vehicles is constant (20 m/s). The ego's initial speed is also 20 m/s, and its action space corresponds to the following longitudinal acceleration values: $a\in \altmathcal{A}=\{-4,-2,-1,0,1,2\}$ m/s$^{2}$. Since the path is already determined (making a left turn), the problem boils down to planning a speed profile for the ego which will complete the task in a minimum amount of time while avoiding catastrophic collisions. A collision is detected when the distance between the ego and target vehicles drops below a threshold value.

To penalize collisions and encourage faster episode termination, the reward is set-up as follows per time step $t$:
\begin{equation}
r_t=
\left\{\begin{matrix}
-5 & \text{if collision} \\
-0.1 & \text{otherwise}
\end{matrix}\right. .
\label{eq:cross_reward_model}
\end{equation}

We concurrently train an optimal agent $\pi^*$, along with one pseudo-agent $\pi_1$. We use deep $Q$-networks \cite{mnih2015}, using double $Q$-learning as well as non-prioritized experience replay \cite{schaul2015prioritized}. The input state to the $Q$-network is the concatenation of the position $x_{ego}$ and speed $\dot{x}_{ego}$ of the ego vehicle, along with the position and computed time-to-collision (ttc) of the 3 nearest targets:

\begin{equation*}
    s=\{x_{ego},\dot{x}_{ego},x_1,\text{ttc}_1,x_2,\text{ttc}_2,x_3,\text{ttc}_3\} .
\end{equation*} 
Each of the components of $s$ are normalized with respect to a maximum value. 

We use the following path-metric $\altmathcal{M}$ between the pseudo-agent, and the optimal agent's memory buffers (in practice we replace the expectation operator by the mean value over the last 100 samples of the agent's memory):

\begin{flalign}
    && \altmathcal{M}(\altmathcal{H}_{\pi_1}, &  \mathbb{E}\left[\altmathcal{H}_{\pi^*}\right]) = \nonumber &&\\
    && & \int \Big \vert \mu\left(\phi\left(\altmathcal{H}_{\pi_1}\right)\right) - \mu\left(\phi\left(\mathbb{E}\left[\altmathcal{H}_{\pi^*}\right]\right)\right) 
    \Big \vert d\phi(s), &&
    \label{eq:metric}
\end{flalign}
where $\phi$ is a state-feature function, and $\mu$ is the density function over state features. In this example we use the speed of the ego vehicle as a state feature $\phi(s)=\dot{x}_{ego}$. The pseudo-reward received at the end of each episode by the corresponding pseudo-agent will be:
\begin{equation}
R^{sub}_{\pi^*}(\altmathcal{H}_{\pi_1}) = -\frac{\alpha}{\altmathcal{M}(\altmathcal{H}_{\pi_1},\mathbb{E}\left[\altmathcal{H}_{\pi^*}\right]) + \delta}    
\label{eq:negative_rwd_implementation}
\end{equation}
with scaling factors:
\begin{center}
    $\alpha = 1,\quad \delta=0.1 \text{  .}$
\end{center}

These determine the relative weight of the pseudo-reward, with respect to the regular reward function $R$. A lower value for $\alpha$ will hardly penalize the pseudo-agent for having a similar state distribution to the reference agent, whereas higher weighting will make the pseudo-agent seek to have a highly different state-space trajectory, disregarding the original objective of the task given by the regular reward function. They are fixed by a rough initial sweep.

\section{RESULTS}
\label{sec:results}

\begin{figure}[H]
    \centering
   \includegraphics[width=8cm]{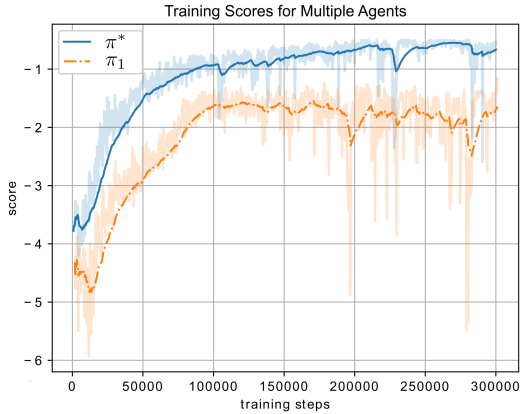}
    \caption{Training scores for both agents. Each is trained for 300k steps. $\pi^*$ is the optimal agent, $\pi_1$ is the pseudo-agent.}
    \label{fig:cross_training_score}
\end{figure}

Fig. \ref{fig:cross_training_score} shows the training scores for both agents. We clearly see the second agent's convergence to its optimal  performance 'lags' behind that of the optimal agent. This is most likely due to the fact that the pseudo-reward term $R^{sub}_{\pi^*}$ depends on the states present in the memory buffer for $\pi^*$.

\begin{figure}[H]
    \centering
    \includegraphics[width=8cm]{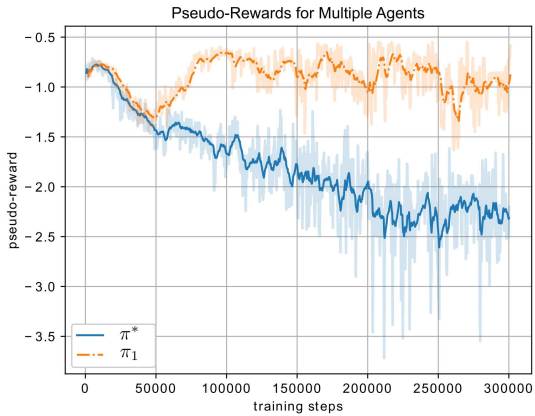}
    \caption{Pseudo-reward, $R^{sub}_{\pi^*}$ calculated for both agents. The values for $\pi^*$ are computed only for comparison to the values used by $\pi_1$.}
    \label{fig:cross_neg_rwd}
\end{figure}

\begin{figure}[H]
    \centering
    \includegraphics[width=8cm]{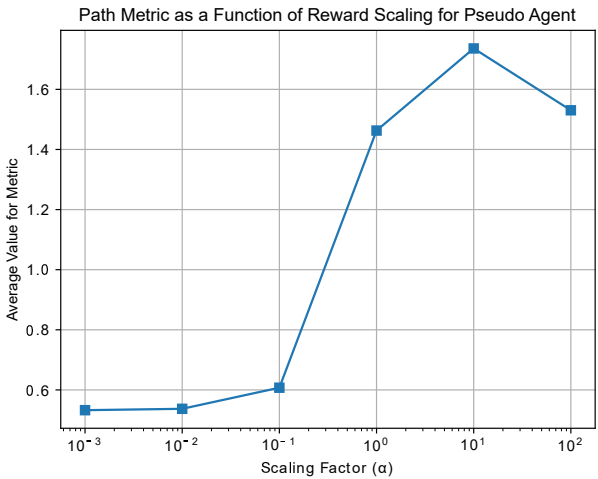}
    \caption{Average values for $\altmathcal{M}$ on the final 100 episodes of each pseudo-agent, for different pseudo-reward scaling $\alpha$.} 
    \label{fig:scaling_alpha}
\end{figure}

\noindent Hence $R^{sub}_{\pi^*}$ cannot be stable until $\pi^*$ has converged, and there is little change in its memory's state distribution. This prevents the corresponding pseudo-agent, $\pi_1$, from converging earlier. Interestingly, $\pi_1$ reaches its best score before $\pi^*$: in our implementation, the optimal path (accelerating before the 1st target vehicle) is harder to find through exploration than the sub-optimal one (passing in-between the target vehicles). Once the pseudo-reward is stable enough to dissuade the pseudo-agent from copying the optimal agent's path, it is faster to converge to its new optimal policy (being the original sub-optimal policy).

\begin{figure*}[t!]
    \centering
\begin{subfigure}{0.32\linewidth}
\includegraphics[width=\linewidth]{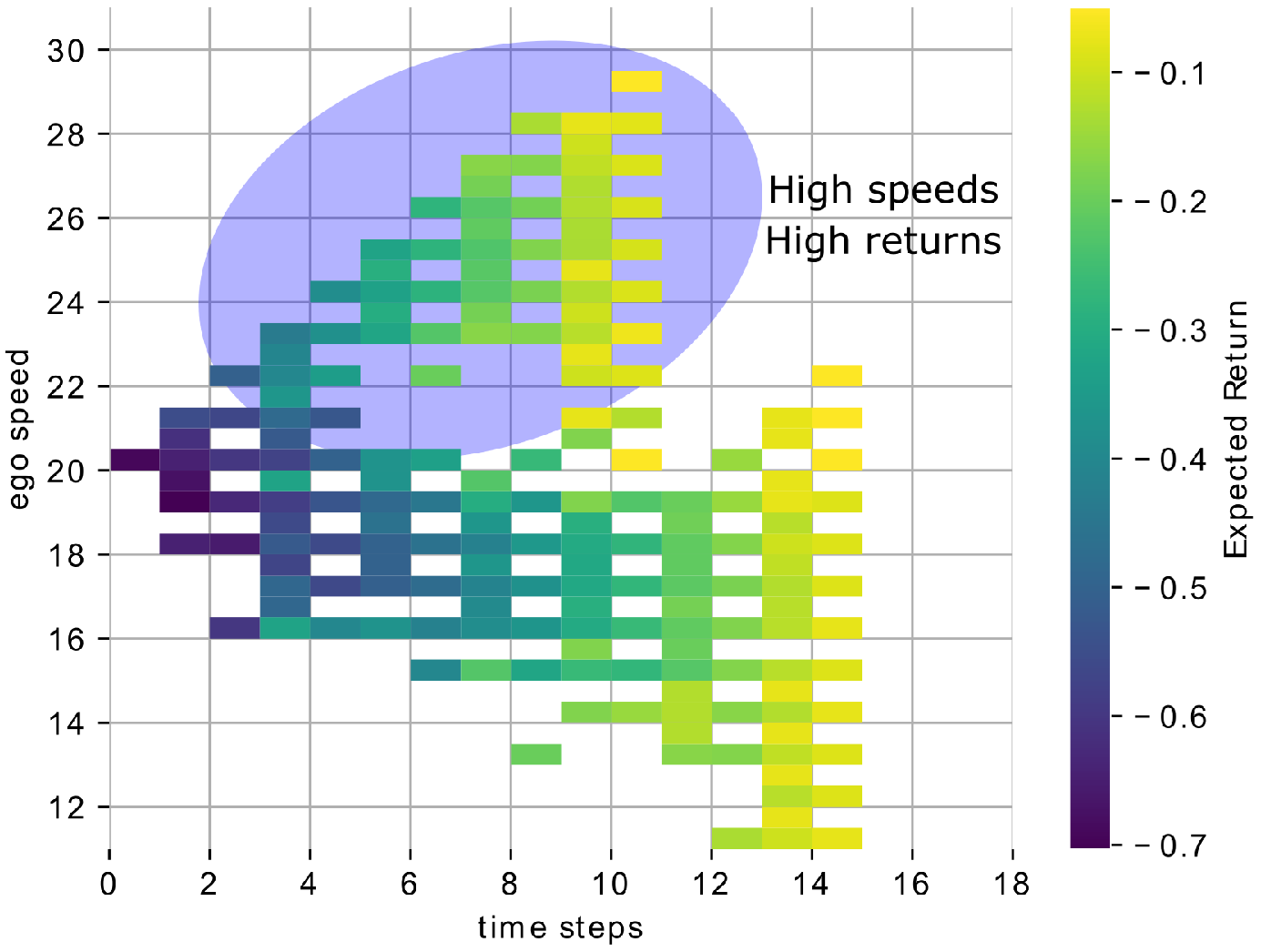}
    \caption{}
    \label{fig:true_q_fn0}
\end{subfigure}
\begin{subfigure}{0.32\linewidth}
\includegraphics[width=\linewidth]{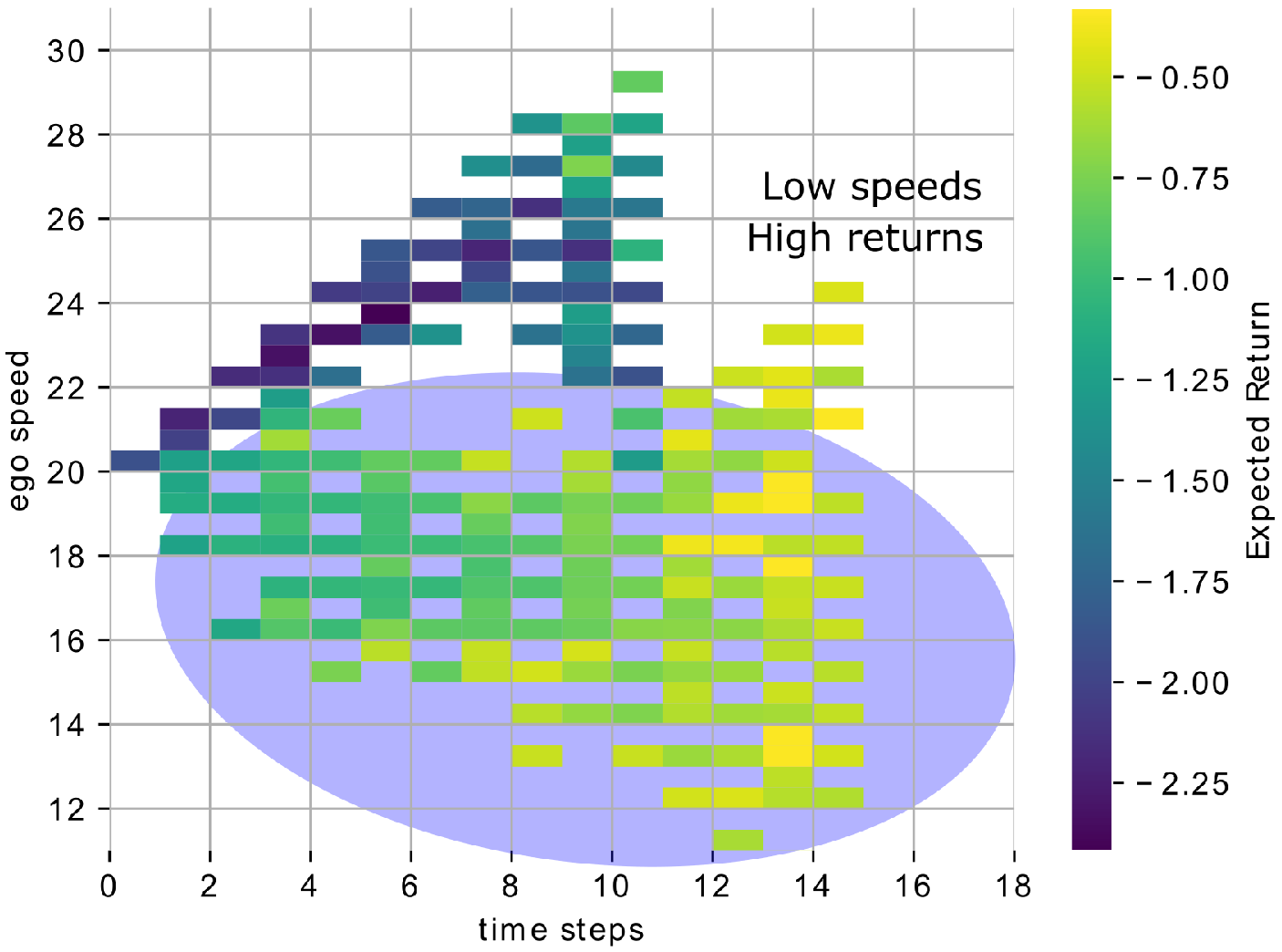}
    \caption{}
    \label{fig:true_q_fn}
\end{subfigure}
\begin{subfigure}{0.32\linewidth}
\includegraphics[width=\linewidth]{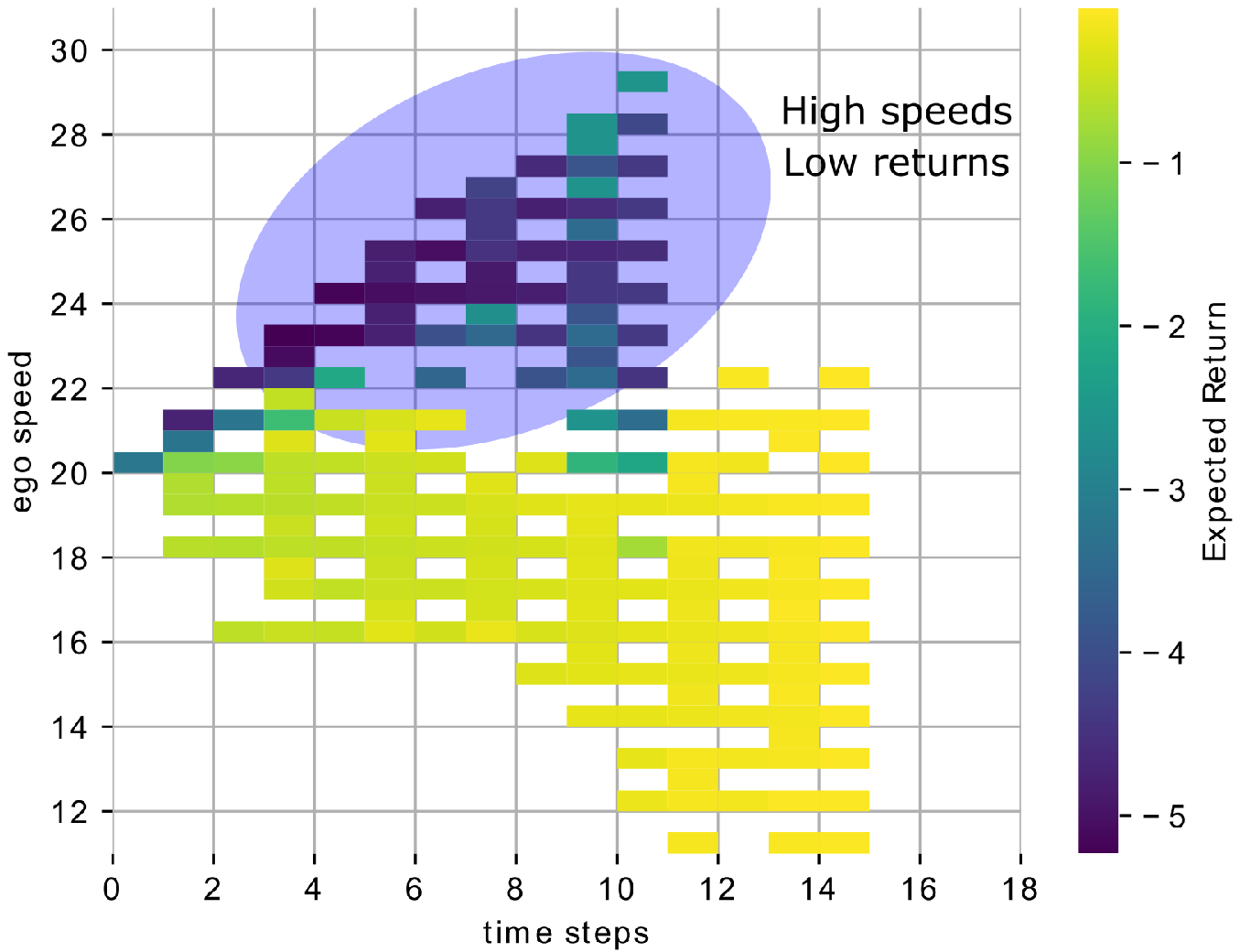}
    \caption{}
    \label{fig:true_q_fn1}
\end{subfigure}
\caption{$Q$-functions evaluated at different areas of feature space: (a) $Q^{\pi^*}$ unaffected by the pseudo-reward, favors higher-speed trajectories. (b) $Q^{\pi_1}$ using pseudo-rewards ($\alpha=1$) favors lower-speed trajectories. (c) $Q^{\pi^*}$ with increased uncertainty on target position, higher-speed trajectories result in a collision with first target vehicle.}
\end{figure*}

Looking at Fig. \ref{fig:cross_neg_rwd}, we see that both agents' path metrics are similar for approximately the first 50k steps, after which their policies start diverging. This means both had similar state distributions (mainly due to a high degree of random exploration) until that point. The value of $R^{sub}_{\pi^*}(\altmathcal{H}_{\pi^*})$ keeps decreasing while the optimal agent is converging to its best policy, and levels out once it converges to its peak performance at approximately 200k steps. $R^{sub}_{\pi^*}(\altmathcal{H}_{\pi_1})$ however levels out quite soon, closely corresponding to $\pi_1$ reaching its peak performance. Though it is still changing due to the changing state distribution in $\pi^*$'s memory buffer, this is hardly seen on the plotted values compared to the random oscillations. 

Fig. \ref{fig:scaling_alpha} shows the effect that modifying the parameter $\alpha$ has on the resulting policy learned by the pseudo-agent $\pi_1$. As mentioned in section \ref{sec:fallback_strategies}, the pseudo-reward must be scaled in such a way to fulfill both conditions (\ref{eq:state_distrib}) and (\ref{eq:adjusted_value_fn}). Learning with pseudo-agents can fail if it is not scaled properly. We see that there is a critical value for $\alpha$, after which the pseudo-agent switches to a sufficiently different behaviour, according to condition (\ref{eq:state_distrib}). In this case, we can deduce that any value for $d$ in the approximate interval $[0.8, 1.4]$ is suitable. Values $< 0.8$ will not steer the pseudo-agent towards a trajectory different to the optimal agent, whereas values $> 1.4$ would falsely rule out policies which we can consider as being heuristically different.

Figs. \ref{fig:true_q_fn0} and \ref{fig:true_q_fn} show the ground truth for $Q^{\pi^*}$ and $Q^{\pi_1}$ respectively, in the ego trajectory feature space, represented as a 2D-tuple of ego vehicle speed, along with the corresponding time step of the episode $(t,\dot{x}_{ego})$. In our use-case, this representation is sufficient to see the difference between varying ego behaviours. We can see in Fig. \ref{fig:true_q_fn0} that with the unmodified reward, the optimal agent $\pi^*$ prefers trajectories having higher speeds, as they correspond to a shorter episode duration which is optimal in the sense of the original reward structure. In Fig. \ref{fig:true_q_fn}, adding an extra pseudo-reward changes the optimization landscape, and tends to steer the pseudo-agent towards areas of lower ego speeds. In all figures, we sampled 10 trajectories from different instances of both $\pi^*$ and $\pi_1$, and plotted the mean $Q$-value for each $(t,\dot{x}_{ego})$ pair.

Fig. \ref{fig:true_q_fn1} shows the change in the expected returns in the case where there is an increase in uncertainty around the first target vehicle's position. In our experiments, we modelled a local increase in sensor uncertainty by increasing the effective collision radius of the first target vehicle by $50\%$. This modification leads to a sharp drop in the performance of $\pi^*$, whereas the state-subspace exploited by $\pi_1$ remains safe and unaffected. We can see that the new optimal policy in the case of Fig. \ref{fig:true_q_fn1}, is also reflected in Fig. \ref{fig:true_q_fn} after adding the pseudo-reward term. This will allow us to use $\pi_1$ as a valid fallback strategy during execution, if ever there is a change in the environment that would not have been accounted for during the initial training phase.

\section{CONCLUSION}
\label{sec:conclusion}
In this paper, we have introduced a new objective in an RL learning pipeline: keeping track of, and learning, sub-optimal policies encountered during the initial training phase.  We have shown that through an intuitive modification of the reward model, that we are able to consistently learn these sub-optimal policies in the case of a driving scenario. 

The context of this work is intended for methods to be applied to model-free problem statements. In the case where the model, even a partial model, or estimation thereof is available to the agent, we gain access to more powerful and data-efficient methods for dealing with introduction of local uncertainties to the MDP.

It is our goal to later combine this work with a hierarchical controller, to be able to quickly switch from optimal to fallback policies in the case of unexpected environment change during the execution phase. This will allow an autonomous vehicle agent to make use of its fallback strategies learned during training, according to its perception of the environment, much like a human would.

\bibliographystyle{IEEEtran}
\bibliography{references}

\addtolength{\textheight}{-12cm}   

\end{document}